\documentclass[lettersize,journal]{IEEEtran}
\usepackage{amsmath,amsfonts}
\usepackage{algorithmic}
\usepackage{algorithm}
\usepackage{array}
\usepackage[caption=false,font=normalsize,labelfont=sf,textfont=sf]{subfig}
\usepackage{textcomp}
\usepackage{stfloats}
\usepackage{url}
\usepackage{verbatim}
\usepackage{graphicx}
\usepackage{cite}
\usepackage{xcolor}

\begin{document}

\title{A Dynamic Transformer Network for Vehicle Detection}

\author{    Chunwei Tian, \IEEEmembership{Member, IEEE}, 
            Kai Liu, 
            Bob Zhang, \emph{Senior Member}, \emph{IEEE},
            Zhixiang Huang, \IEEEmembership{Senior Member, IEEE}, 
            Chia-Wen Lin, \IEEEmembership{Fellow, IEEE}, 
            and David Zhang,  \IEEEmembership{Life Fellow, IEEE}
\thanks{This work was supported in part by National Natural Science Foundation of China under Grant 62201468. (\textit{Corresponding author: Bob Zhang.})}
\thanks{Chunwei Tian and Kai Liu are with the School of Computer Science and Technology, Harbin Institute of Technology, Harbin 15001, China. (Email: chunweitian@hit.edu.cn, liukaiwebsite@163.com)}
\thanks{Bob Zhang is with the PAMI Research Group, University of Macau, 999078, Macao Special Administrative Region of China. (Email: bobzhang@um.edu.mo)}
\thanks{Zhixiang Huang is with the Key Laboratory of Intelligent Computing and Signal Processing, Ministry of Education and Key Laboratory of Electromagnetic Environmental Sensing, Anhui University, Hefei, 230601, China. (Email:zxhuang@ahu.edu.cn)}
\thanks{Chia-Wen Lin is with the Department of Electrical Engineering and the Institute of Communications Engineering, National Tsing Hua University, Hsinchu, Taiwan. (Email: cwlin@ee.nthu.edu.tw)}
\thanks{David Zhang is with the School of Data Science, The Chinese University of Hong Kong (Shenzhen), Shenzhen, 518172, China. (Email:davidzhang@cuhk.edu.cn)}}

\markboth{
}%
{Shell \MakeLowercase{\textit{et al.}}: A Sample Article Using IEEEtran.cls for IEEE Journals}


\maketitle

\begin{abstract}
Stable consumer electronic systems can assist traffic better. Good traffic consumer electronic systems require 
collaborative work between traffic algorithms and hardware. However, performance of popular 
traffic algorithms containing vehicle detection methods based on deep networks via learning data relation rather than learning differences in different lighting and occlusions is limited.  In this paper, we present a dynamic Transformer network for vehicle detection (DTNet). DTNet utilizes a dynamic convolution to guide a deep network to dynamically generate weights to enhance adaptability of an obtained detector. Taking into relations of different information account, a mixed attention mechanism based channel attention and Transformer is exploited to strengthen relations of channels and pixels to extract more salient information for vehicle detection. To overcome the drawback of difference in an image account,  a translation- variant convolution relies on spatial location information to refine obtained structural information for vehicle detection. Experimental results illustrate that our DTNet is competitive for vehicle detection. Code of the proposed DTNet can be obtained at \url{https://github.com/hellloxiaotian/DTNet}.
\end{abstract}

\begin{IEEEkeywords}
Dynamic convolution, Transformer, Hybrid Attention Mechanism, Vehicle Detection.
\end{IEEEkeywords}

\section{Introduction}
\label{sec:introduction}
\IEEEPARstart{W}{ith} the development of digital devices and artificial intelligence techniques, more consumer electronic systems have been 
deployed on popular consumer electronic devices, i.e., drones, cameras and smart phones, which have been widely applied to traffic detection \cite{singh2024smart,gao2024monoli}. However, complex traffic conditions may have higher requirements for 
robust traffic detection algorithms, i.e., vehicle detection methods. With the development of surveillance cameras and computer vision technology, detection techniques have been developed \cite{dalal2005histograms}. For instance, Dalal et al. \cite{dalal2005histograms} designed a two-stage detector for object detection. That is, they calculated the gradient histograms in local regions as inputs of a support vector machine model to detect objects, which can make a trade-off between detection accuracy and computational speed. \textcolor{black}{Taking the robustness of vehicle detection into account, Chen et al. \cite{chen2012vehicle} adopted a new background Gaussian Mixture Model and shadow removal method to handle sudden changes in illumination and camera vibrations.} In scenarios of limited computational resources, Sara et al. \cite{minaeian2015vision} leveraged the information of key points between frames to effectively guide and enhance the optical flow algorithm for motion target detection. These detection methods with good performance are extended for vehicle detection. Taking a three-dimensional position and pose information of vehicles into account, Karaimer et al. \cite{karaimer2017detection} proposed to project and register silhouettes of each planar rigid body to improve the accuracy of vehicle detection. To enhance the adaptability of vehicle detection in different scenes, Chang et al. \cite{chang2009online} embedded a training strategy of dynamically selecting the most representative samples into the AdaBoost classifier to improve the robustness of vehicle detection. To improve detection accuracy, Zhou et al. \cite{zhou2007moving} utilized motion trajectory information of vehicles to filter out some false and repeated detection results. Although these methods can achieve good performance in certain scenarios for vehicle detection, they may require manual setting parameters and complex optimization algorithms. That would cause big computational costs. 

Convolutional neural networks (CNNs) with strong feature extraction functions are extended to object and vehicle detection \cite{lim2017real}. For instance, 
Duan et al. \cite{duan2019centernet} converted object detection questions to detection questions of center and key points in terms of performance and efficiency in object detection. Motivated by that, similar techniques can be used for vehicle detection. Rani et al. \cite{rani2021littleyolo} used spatial pyramid and multi-scale techniques to improve YOLOv3 to achieve a lightweight vehicle detector. To improve detection accuracy, Hu et al. \cite{hu2018sinet} combined context perceptual pooling and multi-branch decision network to refine structural information and reduce the distance of intra-class features for vehicle detection. Also, Song et al. \cite{song2019vision} used segmentation algorithms to divide into two parts, i.e., remote and proximal areas, and YOLOv3 to respectively detect them to improve the accuracy of small vehicle detection. \textcolor{black}{Dong et al. \cite{dong2022lightweight} utilized an attention mechanism to suppress non-critical information, and introduced the Ghost module to achieve a balance between vehicle detection performance and efficiency. Roy et al. \cite{roy2022multi} analyzed the shortcomings of the visual modality and introduced data from other modalities, complementing visual information to enhance the performance of vehicle detection. To enhance the robustness of vehicle detection in different scenarios, Zhang et al. \cite{zhang2022real} utilized the Flip-Mosaic algorithm to improve the perception of small targets and reduce the false detection rate of vehicles.} Although these methods can detect vehicles, their flexibility may get poor for complex traffic scenes, i.e., dim lighting.  

In this paper, we propose a dynamic Transformer network for vehicle detection (DTNet). DTNet uses a dynamic convolution to dynamically learn parameters to improve the robustness of an obtained detector, according to different scenes. Taking into relations 
between channels and pixels account, a mixed attention mechanism is used to mine salient information to improve performance in vehicle detection. To overcome differences drawback within an image, a variant utilizes spatial location information to refine more structural information for vehicle detection. Experimental results show that our DTNet is superior to vehicle detection. 

The contributions of this paper can be conducted as follows. 

(1)	A dynamic convolution is applied to dynamically generate parameters to enhance the robustness of an obtained classifier for vehicle detection in different scenes, according to different traffic scenes. 

(2)	 A mixed attention mechanism relies on enhancing relations of channels and 
pixels to extract more salient information in vehicle detection. 

(3)	 A translation variant convolution uses spatial location information to refine obtained information to improve accuracy in vehicle detection.   

The remaining parts of this paper can be listed as follow. Section II provides related work on deep CNNs and dynamic networks for image applications. Section III describes the proposed object detection method. Section IV gives datasets, experimental settings, and experimental results. Section V concludes this paper.

\section{Related work}
\subsection{Deep CNNs for object detection}
Due to strong expressive abilities, deep CNNs are extended to extended for vehicle detection \cite{lim2017real}. Specifically, vehicle detection algorithms based on CNNs can be divided into two kinds, i.e., two- and one-stage algorithms. 

A two-stage CNN for vehicle detection uses a CNN to find the location of objects and then utilizes another CNN to recognize these objects in the areas. Mo et al. \cite{mo2020improved} combined feature amplification and oversampling data augmentation to enhance Faster R-CNN, which enhanced feature extraction ability of hidden information and overcomed negative influence of sample imbalance to improve performance of vehicle detection. Manana et al. \cite{manana2018preprocessed} exploited Sobel edge operator and Hough Transform to detect lane regions and then utilized a faster R-CNN to classify these vehicles in obtained regions, which can improve the accuracy of vehicle detection. 
Although these methods have obtained good performance in vehicle detection, their detection speed is limited. To overcome the drawback, one-stage CNN for vehicle detection is developed \cite{balamuralidhar2021multeye}.

One-stage CNN for vehicle detection directly predicts regions and categories of objects. For instance, Balamuralidhar et al. \cite{balamuralidhar2021multeye} introduced a semantic segmentation head into a CNN to build a multi-task training framework to enhance the robustness of a vehicle detector. To deal with negative effect of scale variation and occlusion of objects, Kim et al. \cite{kim2018performance} referred to spatial pyramid pooling techniques into YOLOv3 to obtain multi-scale features to improve the accuracy of vehicle detection. \textcolor{black}{
Kumar et al.\cite{kumar2022deep} combined temporal information with the YOLO architecture to achieve the detection of both light and heavy vehicles. To adapt different weather conditions, Humayun et al. \cite{humayun2022traffic} employed dataset augmentation techniques during the training process of the YOLO architecture by adjusting hue, saturation, exposure to enhance the performance of vehicle detection.} To improve the efficiency of vehicle detection, a Ghost module was fused into a YOLOv5 \cite{dong2022lightweight}. Also, an attention module is introduced to select vital information to improve the detection accuracy of vehicle detection \cite{dong2022lightweight}. \textcolor{black}{To reduce the computational load of convolutional kernel operations, Charouh et al. \cite{charouh2022resource} utilized preprocessing techniques to select smaller mobile regions as inputs for a single-stage vehicle detector, e.g., YOLOv5s to improve its efficiency.} According to the mentioned illustrations, we design an adaptive network for vehicle detection based on the one-stage algorithm in terms of detection performance and efficiency. To deal with different traffic scenes, we can dynamically adjust parameters to enhance the adaptability of the vehicle detection model for various traffic conditions, such as different lighting, occlusions, and weather variations, according to different input vehicle images in this paper. 

\subsection{Dynamic network for image applications}
CNN models usually are independent of test input data \cite{mo2020improved}, which may cause performance degradation in image applications in real scenes. To tackle this problem, dynamic convolutions are developed \cite{chen2020dynamic}. That is, they can dynamically adjust parameters to adapt different input images to improve the performance of image applications. That is, Chen et al. \cite{chen2020dynamic} combined convolutional kernels to dynamically generate parameters to improve the robustness of a classifier. To reduce computational costs, Thomas et al. \cite{verelst2020dynamic} dynamically aggregate significant pixels to make a trade-off between performance and computational resources in image recognition. Taking into differences of different regions account, Chen et al. \cite{chen2021dynamic} dynamically assign different convolutional kernels to different regions to extract region-sensitive information, which can improve classification results. To extract salient information, Hu et al. \cite{hu2018squeeze} utilized an attention mechanism to adaptively adjust effects of channels to promote classification performance. Inspired by that, we employ a dynamic idea to guide a CNN
to achieve a network for vehicle detection in terms of different
traffic scenes in this paper.

\begin{figure*}[!htbp]
\centerline{\includegraphics[width=5in]{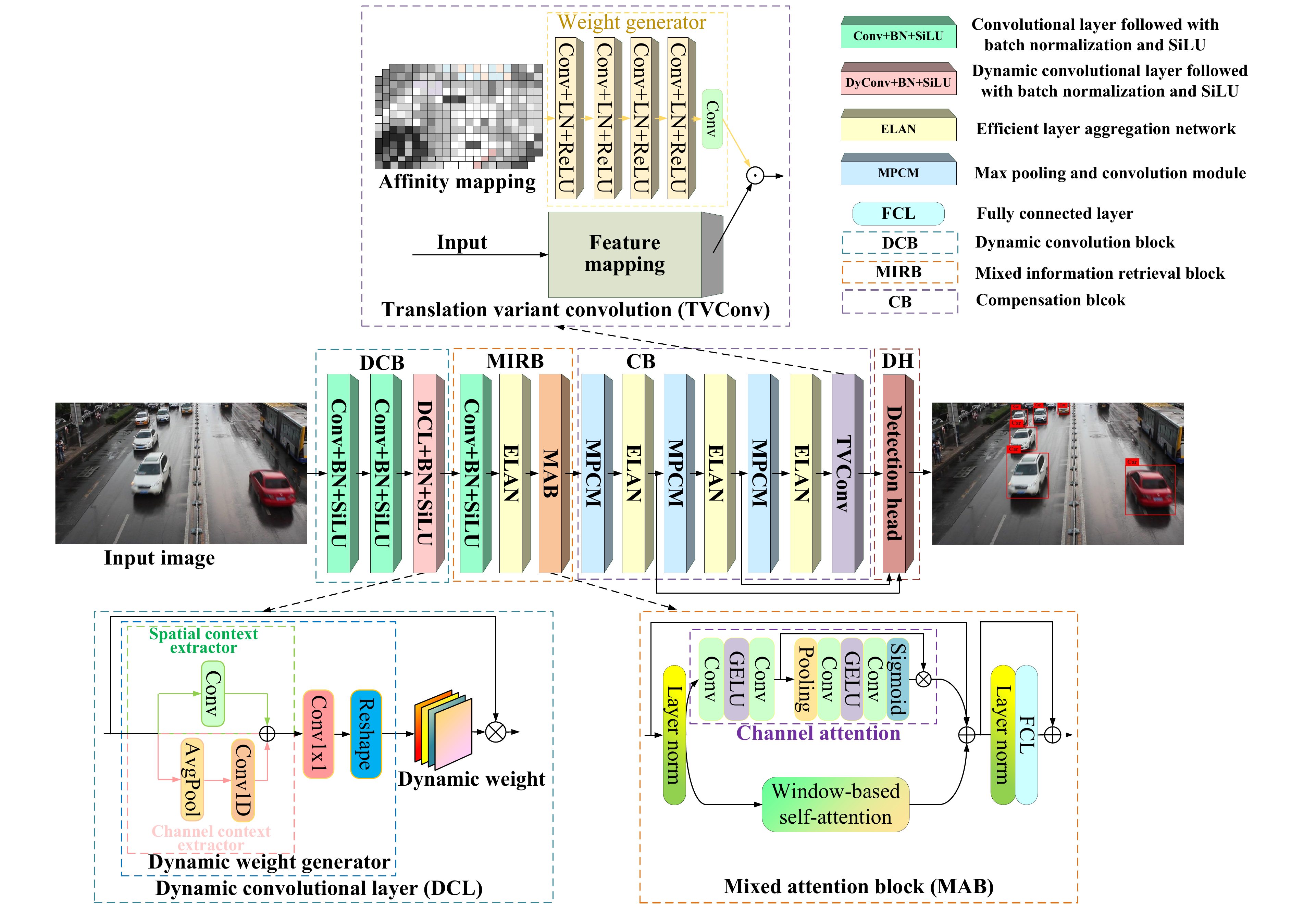}}
\caption{\textcolor{black}{Architecture of DTNet.}}
\label{fig1}
\end{figure*}

\section{Proposed Method}
\subsection{Network architecture}
\textcolor{black}{Vehicle detection methods are effective in certain traffic scenes in general. However, the traffic environment is affected by illuminations and occlusions.} To address these issues, we present a translation variant network in vehicle detection as well as DTNet as shown in Fig.1. \textcolor{black}{According to Refs. \cite{zhang2020residual}, we choose a modular way to express the proposed method.} That is, DTNet is composed of a dynamic convolutional block (DCB), a mixed information retrieval block (MIRB), compensation block (CB) and detection head (DH). DCB is used to dynamically adjust parameters to enhance the robustness of an obtained classifier for different scenes. To extract more salient information, MIRB is used to simultaneously mine complementary information, according to relations between channels and pixels. Due to differences within an image, CB is applied to refine obtained information to improve the effect of vehicle detection, according to spatial location information. Finally, a DH is used to predict locations and categories of vehicles. Mentioned illustrations can be converted as Eq. (1). 
\begin{equation}
\begin{array}{ll}
l,c = DTNet(I)\\
{\rm{       = }}DH{\rm{(}}CB{\rm{(}}MIRB{\rm{(}}DCB{\rm{(}}I{\rm{))))}}
\end{array}
\end{equation}

\noindent{where $I$ stands for an input image. $DTNet$, $DCB$, $MIRB$, $CB$, and $DH$ strand for functions of DTNet, DCB, MIRB, CB, and DH, respectively. $l$ is used to represent the location of an object. $c$ is a category of recognized object. \textcolor{black}{DTNet can be trained by a loss function \cite{wang2023yolov7}.}}

\subsection{Dynamic convolution block}
\textcolor{black}{CNN models with fixed weights usually deal with vehicle detection. Although they have obtained good performance in vehicle detection, their performance may decrease for complex scenes. Also, dynamic convolutions \cite{shen2023adaptive} can adjust parameters to improve robustness of a CNN model, according to different scenes. Inspired by that, we fuse a dynamic convolution as a dynamic convolution block into a CNN for vehicle detection.} That is, dynamic convolution block is composed of two stacked Conv+BN+SiLU and a DCL+BN+SiLU, where Conv+BN+SiLU is a combination of a convolutional layer, batch normalization (BN) \cite{ioffe2015batch} and SiLU \cite{elfwing2018sigmoid}. That can extract structural information from given original images. DCL+BN+SiLU is combination of a dynamic convolutional layer, BN and SiLU. That can dynamically update parameters to improve robustness of an obtained detector, according to different traffic scenes. To visually show the procedure, DCB can be conducted as the following equation. 

\begin{equation}
\begin{array}{ll}
{O_{DCB}} &= DCB(I)\\
 &= DBS(CBS(CBS(I)))
\end{array}
\end{equation}
where $DCB$ is a function of DCB. $CBS$ and $DBS$ stand for functions of Conv+BN+SiLU and DCL+BN+SiLU, respectively. ${O_{DCB}}$ expresses an output of DCB. More detailed information of DCL \cite{shen2023adaptive} can be introduced as follows. 

DCL includes parallel and serial components. The parallel component is composed of a spatial context extractor (SCE) including a convolutional layer with $3\times3$  and a channel context extractor (CCE) including an average pooling and a convolutional layer with $1\times3$. They can complement context information with different dimensions. These obtained information can be fused by a residual learning mechanism ($ \oplus$) in Fig.1. A convolutional layer with $1\times1$ is used to refine these obtained information. To make obtained information have same size with an input of DCL, a reshape operation is used behind a convolutional layer with $1\times1$ to obtain as dynamic weights. Obtained dynamic weights multiply an input of DCL to make an obtained detector adapt varying traffic scenes. DCL can be formulated as Eq. (3). 

\begin{equation}
\begin{array}{ll}
{O_{DCL}}({O_t}) &= {C_1}(SCE({O_t}) + CCE({O_t})) \times {O_t}\\
& = {C_1}(C({O_t}) + {C^{'}}(Avg({O_t}))) \times {O_t}
\end{array}
\end{equation}

\noindent{\textcolor{black}{where $O_{t}$ is used to denote an input of DCL as well as the output of cascaded CBS, and $O_{DCL}$ stands for the output of DCL, which acts MIRB.} $SCE$ and $CCE$ are used to express functions of SCE and CCE, respectively. $C_1$, $C$ and $C^{'}$ are exploited to denote a convolutional layer with $1\times1$, a convolutional layer with $3\times3$, a convolutional layer with $1\times3$, respectively. $Avg$ is an average pooling operation. $+$ is a residual learning operation. $\times$ is a multiplication as well as $\otimes $ in Fig.1.} 

\subsection{Mixed information retrieval block}
\textcolor{black}{To extract more salient information, a mixed information retrieval block is used for vehicle detection.} It mainly includes a Conv+BN+SiLU, an efficient layer aggregation network (ELAN) and a mixed attention block (MAB) in Fig.1. A Conv+BN+SiLU is used to refine obtained information from the DCB. ELAN \cite{wang2023yolov7} is exploited to extract more accurate structural information. To extract more salient information, a MAB \cite{chen2023activating} is used to mine complementary structural information for vehicle detection. Its procedure can be formulated as follows. 

\begin{equation}
\begin{array}{ll}
{O_{MIRB}} &= MIRB({O_{DCB}})\\
 & = MAB(ELAN(CBS({O_{DCB}})))
\end{array}
\end{equation}

\noindent where $O_{MIRB}$ is an output of MIRB. $MIRB$, $CBS$, $ELAN$ and $MAB$ are functions of MIRB, CBS, ELAN and MAB, respectively. Specifically, MAB \cite{chen2023activating} is composed of two different attention mechanisms, i.e., a channel attention mechanism and a window-based self-attention mechanism. A channel attention can enhance relations of channels to extract salient information. Window-based self-attention mechanism can enhance effects of pixels to extract salient information. They are complementary for vehicle detection. More detailed information of MAB can be obtained from Ref. \cite{chen2023activating}.

\subsection {Compensation module}
\textcolor{black}{Taking differences within an image into account, a compensation block is proposed. It consists of a max pooling and convolution module (MPCM) as the first, third and fifth layers, ELAN as the second, fourth, sixth layers, and a translation variant convolution (TVConv). MPCM is used to capture multi-scale information to improve performance of vehicle detection. ELAN is used to extract more accurate structural information. Also, TVConv \cite{chen2022tvconv} utilizes spatial location information to refine obtained information to improve recognition rate in vehicle detection.} The mentioned illustrations can be conducted as follows.  
\begin{equation}
\begin{array}{ll}
{O_{CB}} &= CB({O_{MIRB}})\\
 &= TVConv(ELAN(MPCM(ELAN(MPCM\\&(ELAN(MPCM({O_{MIRB}}){\rm{))))))}}
\end{array}
\end{equation}
where $O_{CB}$ is an output of CB. $MPCM$, $ELAN$, and $TVConv$ represent functions of MPCM, ELAN, and TVConv, respectively. Specifically, TVConv is implemented via a convolutional operation to act between a weight generator and feature mapping. The weight generator can be composed of four stacked Conv+LN+ReLU and a Conv, where Conv+LN+ReLU is a combination of a convolutional layer, LN and ReLU. All convolutional kernel sizes in the weight generator are $3\times3$. The procedure can be formulated as Eq. (6).

\begin{equation}
\begin{array}{ll}
{O_{CB}} &= TVConv({O_{t1}})\\
&= {O_{t1}} \odot Conv(CLR(CLR(CLR(CLR(A)))))
\end{array}
\end{equation}
where $O_{t1}$ is an input of TVConv, $A$ denotes a learnable affine mapping, $CLR$ stands for a combination of a convolutional layer, a LN and a ReLU. $\odot$ expresses a convolutional function. A detection head (DH)\cite{wang2023yolov7} is used to predict locations and categories of vehicles. 

\section{Experiments}
\subsection{Datasets}
To validate performance of the proposed method for vehicle detection, UA-DETRAC dataset \cite{wen2020ua}, Berkeley Deep Drive (BDD) 100k object detection dataset \cite{yu2020bdd100k} and BIT-Vehicle Dataset \cite{dong2015vehicle} can be chosen to train and test a DTNet model. Also, mean average precision (mAP) is used as metric to test performance for vehicle detection.   

\textbf{UA-DETRAC dataset} \cite{wen2020ua} captures 100 videos at a rate of 25 frames per second. Specifically, 60 video sequences are used as a training dataset and 40 video sequences are used as a test dataset. These videos collected on four vehicle categories with different scales and poses, i.e., cars, buses, vans, and others from different conditions, cloudy, night, sunny and rainy. To accelerate the training process, a training video every 10th frame is selected to construct a training sub-dataset. Also, the number of training frames is 8,639 and the number of test frames is 2,231. 

\begin{figure*}[!htbp]
\centerline{\includegraphics[width=6in]{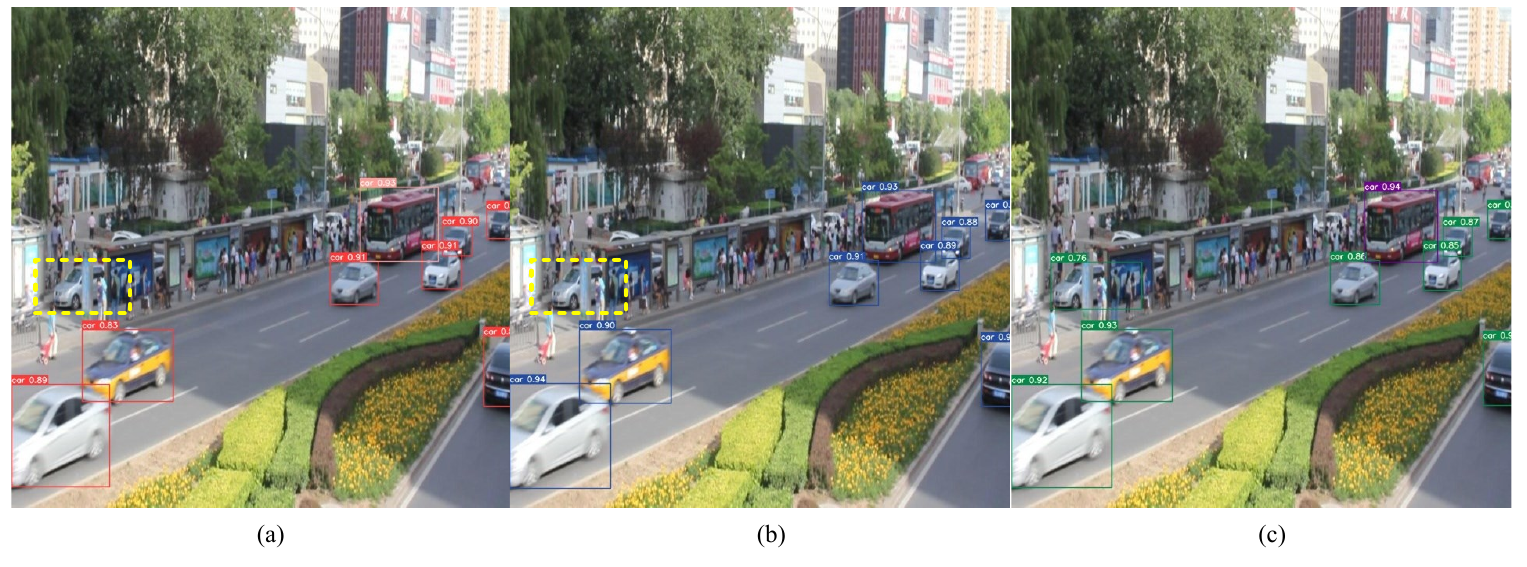}}
\caption{\textcolor{black}{Visual effects of three detection methods on a scene from the UA-DETRAC: (a) YOLOv5s, (b) YOLOv7 and (c) DTNet (Ours).}} 
\label{fig2}
\end{figure*}

\begin{figure*}[!htbp]
\centerline{\includegraphics[width=6in]{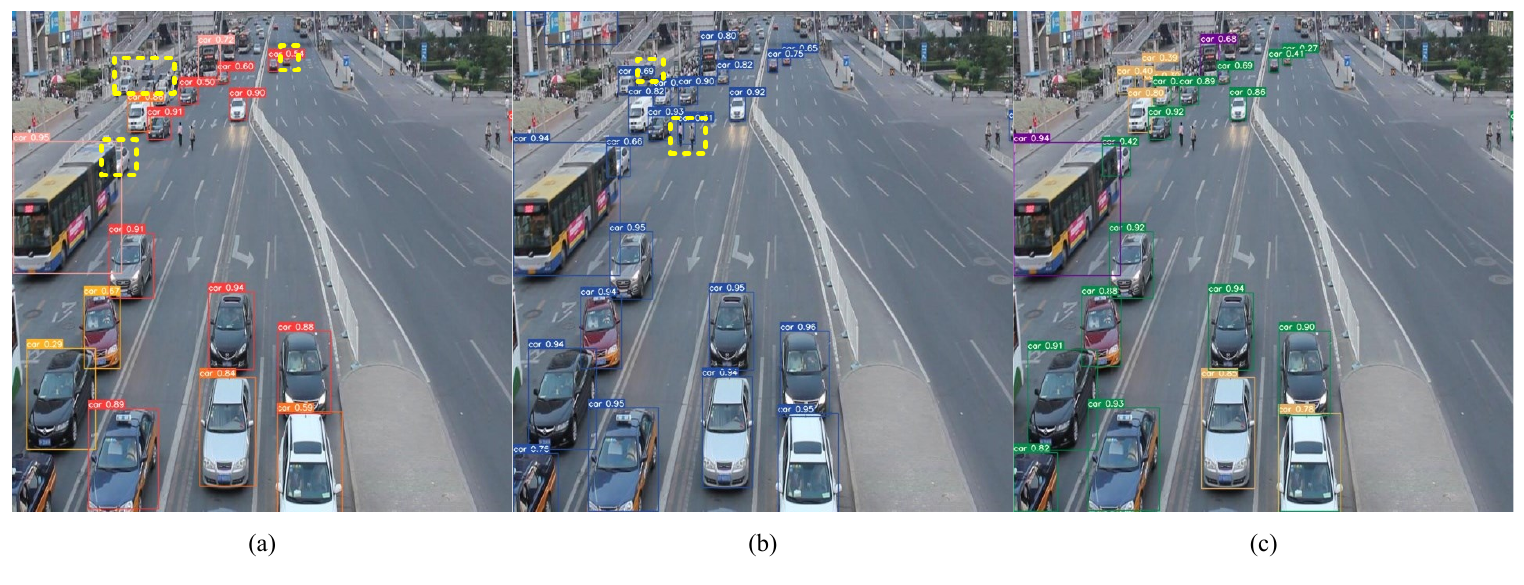}}
\caption{\textcolor{black}{Visual effects of three detection methods on a scene from the UA-DETRAC: (a) YOLOv5s, (b) YOLOv7 and (c) DTNet (Ours).}}
\label{fig3}
\end{figure*}

\textbf{BDD100K dataset} \cite{yu2020bdd100k} consists of 70,000 training frames and 10,000 testing frames, which are used to evaluate performance of vehicle detection in autonomous driving. Training and test datasets are composed of objects with ten types, i.e., Bus, Light, Sign, Person, Bike, Truck, Motor, Car, Train and Rider. Also, they include six scenes, i.e., residential areas, highways, city streets, parking lots, gas stations, and tunnels. Besides, it includes various weather conditions, i.e., sunny, cloudy, overcast, rainy, snowy and foggy. 

\textbf{BIT-Vehicle dataset}  \cite{dong2015vehicle} consists of six different vehicles, i.e., sedan, sport-utility vehicle (SUV), microbus, truck, bus, and minivan. And a training dataset contains 7,880 frames. A test dataset includes 1,970 frames. 

\subsection{Experimental settings }
To obtain a robust detector, we set the batch size and the number of training epochs to 8 and 250, respectively. And we initialized the learning rate to 0.01 and applied the one cycle policy to update the learning rate. We used Stochastic Gradient Descent (SGD) \cite{bottou2012stochastic} to optimize our detector, with a momentum value of 0.937. Other experimental parameters can be set as Ref. \cite{wang2023yolov7}. We use PyTorch of 1.10.2 \cite{stevens2020deep} and Python of 3.8.5. 

We conduct experiments in this papers on Ubuntu of 20.04 with setting of AMD EPYC of 7502P/3.35GHz and CPU of 33 cores. 
Its memory is 128G. Also, it depends on a GPU of a Nvidia GeForce GTX 3090 to improve training speed. Additionally, Nvidia CUDA of 11.1 and cuDNN of 8.04 can improve work efficiency of GPU above.

\begin{figure*}[!htbp]
\centerline{\includegraphics[width=6in]{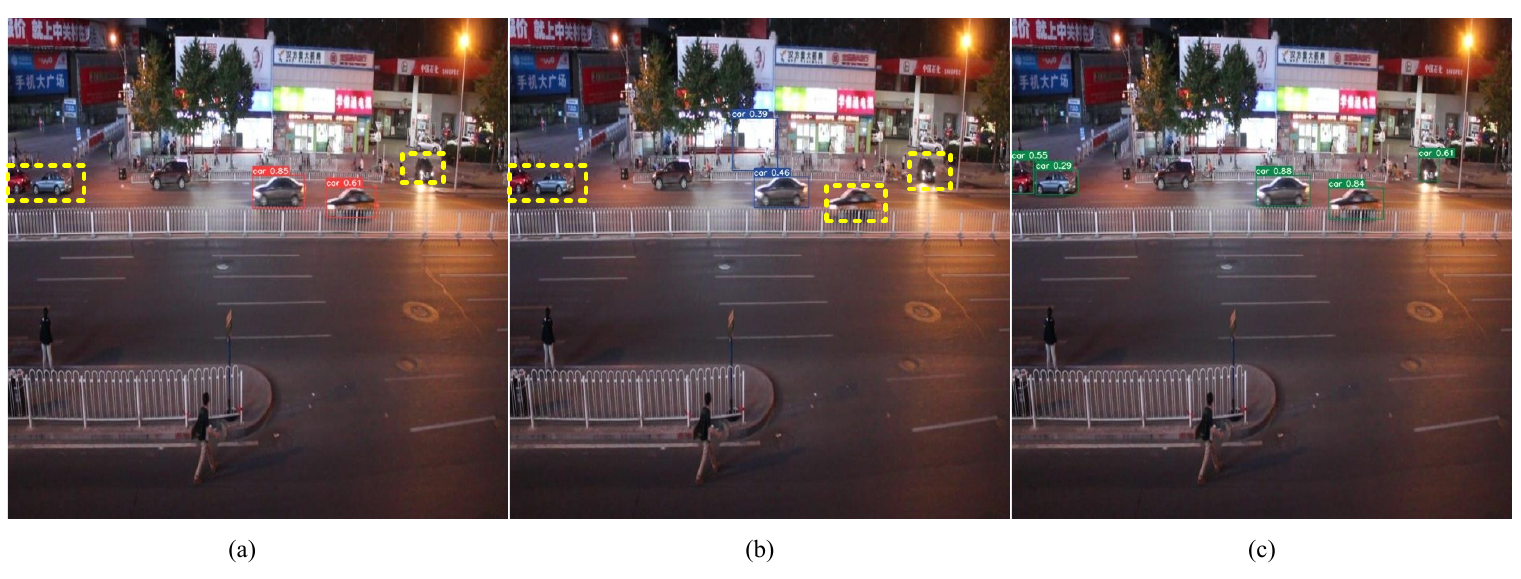}}
\caption{\textcolor{black}{Visual effects of different detection methods on a scene from the UA-DETRAC: (a) YOLOv5s, (b) YOLOv7, and (c) DTNet (Ours).}}
\label{fig4}
\end{figure*}

\begin{table}
\caption{mAP of serval methods on the BIT vehicle dataset.}
\label{table1}
\setlength{\tabcolsep}{3pt}
\centering
\begin{tabular}{|c|c|}
\hline
Methods & mAP \\
\hline
DTNet without DCL, \textcolor{black}{MAB} and \textcolor{black}{TVConv} & 77.1\% \\
DTNet without \textcolor{black}{MAB} and \textcolor{black}{TVConv} & 79.2\% \\
DTNet without \textcolor{black}{TVConv} & 84.5\% \\
DTNet (Ours) & 84.7\% \\
\hline
\end{tabular}
\end{table}

\begin{table}
\caption{Performance of different methods on the BDD100K dataset.}
\label{table2}
\setlength{\tabcolsep}{3pt}
\centering
\begin{tabular}{|c|c|}
\hline
Methods & mAP \\
\hline
YOLOv3 & 25.8\% \\
SSD & 33.9\% \\
WLOD & 34.3\% \\
YOLOv4 & 50.1\% \\
Modified \textcolor{black}{YOLOv4} & \textcolor{blue}{52.7\%} \\
YOLOv5s & 50.6\% \\
AFFB\_YOLOv5s & 51.5\% \\
\textcolor{black}{YOLOv5\_ACmix} & 51.8\% \\
DTNet (Ours) & \textcolor{red}{57.3\%} \\
\hline
\end{tabular}
\end{table}

\begin{table}
\caption{Performance of different methods on the UA-DETRAC dataset.}
\label{table3}
\setlength{\tabcolsep}{3pt}
\centering
\begin{tabular}{|c|c|}
\hline
Methods & mAP \\
\hline
Densenet & 42.0\% \\
RetinaNet & 55.2\% \\
EfficientNet & 60.6\% \\
YOLOv3 & 59.1\% \\
YOLOv4 & 60.0\% \\
YOLOv5s & 55.1\% \\
YOLOv7 & 67.3\% \\
CSIM & 64.5\% \\
\textcolor{black}{DRConv} & 67.4\% \\
\textcolor{black}{DC} & \textcolor{blue}{67.7}\% \\
DTNet (Ours) & \textcolor{red}{69.0\%} \\
\hline
\end{tabular}
\end{table}

\begin{table}
\caption{Performance of different methods on the BIT vehicle dataset.}
\label{table4}
\setlength{\tabcolsep}{3pt}
\centering
\begin{tabular}{|c|c|}
\hline
Methods & mAP \\
\hline
YOLOv3 & \textcolor{blue}{81.3\%} \\
\textcolor{black}{YOLOv4\_AF} & 77.08\% \\
YOLOv5s & 79.7\% \\
YOLOv7 & 77.0\% \\
DTNet (Ours) & \textcolor{red}{84.7\%} \\
\hline
\end{tabular}
\end{table}

\subsection{Ablation studies}
To deal with more complex traffic scenes, we propose a vehicle detection with a translation-variant CNN as well as DTNet. DTNet is based on a YOLOv7, which is divided into four parts, i.e., the first, second, third and fourth parts. The first part fuses a dynamic convolutional layer to be composed of a dynamic convolutional block to dynamically learn parameters to adapt to complex scenes. Its more detailed information can be shown in Section III. B. Its effectiveness can be proved that ‘DTNet without MAB and TVConv’ has obtained higher mAP value than that of ‘DTNet without DCL, MAB and TVConv’ as shown in TABLE I. To extract more salient information, a mixed attention mechanism is gathered into the second part as well as a mixed information retrieval block. Its detailed information can be shown in Section III. C. Its effectiveness can be verified through comparing ‘DTNet’ and ‘DTNet without TVConv’ in terms of mAP in TABLE I. To overcome differences within an image, a translation variant convolutional layer is fused into the third part to implement a compensation block, which can refine obtained information to improve effect of vehicle detection, according to spatial location information. Its effectiveness can be proved that ‘DTNet without MAB’ has obtained a higher mAP value than that of ‘DTNet without MAB and TVConv’ as shown in TABLE I. Finally, DH is used to predict locations and categories of vehicles.

\begin{table}
\caption{\textcolor{black}{Precision, recall}, mAP and \textcolor{black}{mAP with varying thresholds of different methods} on the UA-DETRAC dataset.}
\label{table5}
\setlength{\tabcolsep}{3pt}
\centering
\begin{tabular}{|c|c|c|c|c|}
\hline
Methods & \textcolor{black}{Precision} & \textcolor{black}{Recall} & mAP & \textcolor{black}{mAP with varying thresholds} \\
\hline
YOLOv5s & 63.0\% & 52.1\% & 55.3\% & 47.2\% \\
YOLOv7 & \textcolor{blue}{64.0\%} & \textcolor{blue}{62.6\%} & \textcolor{blue}{65.3\%} & \textcolor{blue}{48.4\%} \\
DTNet (Ours) & \textcolor{red}{64.6\%} & \textcolor{red}{70.3\%} & \textcolor{red}{69.0\%} & \textcolor{red}{51.4\%} \\
\hline
\end{tabular}
\end{table}

\begin{figure*}[!htbp]
\centerline{\includegraphics[width=6in]{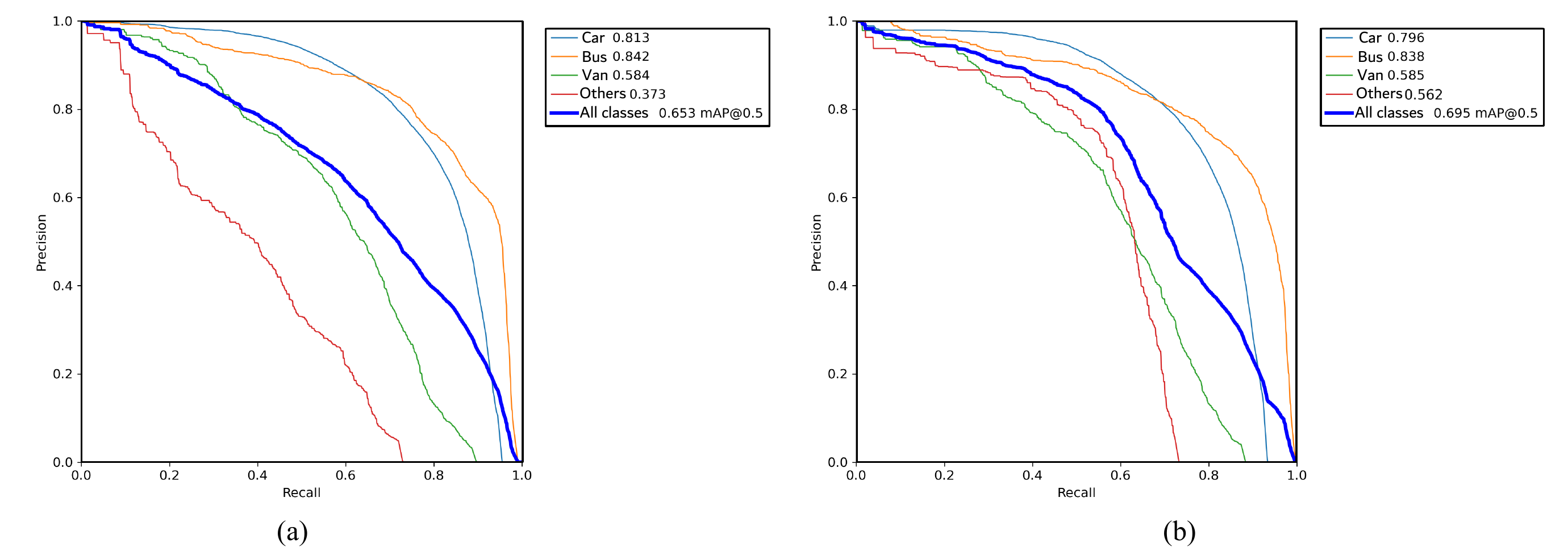}}
\caption{\textcolor{black}{The PR curve of of different detection methods on the UA-DETRAC: (a) YOLOv7 and (b) DTNet (Ours).}}
\label{fig5}
\end{figure*}

\subsection{Comparisons with state-of-the-art vehicle detection methods}
To fairly test detection performance of our DTNet,  quantitative and qualitative methods are chosen to analyze its effect in vehicle detection. Quantitative analysis is that popular methods, i.e., YOLOv3 \cite{redmon2018yolov3}, SSD \cite{liu2016ssd}, Wasserstein Loss based Model for Object Detection (WLOD) \cite{han2020wasserstein}, YOLOv4 \cite{bochkovskiy2020yolov4} and Modified YOLOv4 \cite{li2020deep}, \textcolor{black}{YOLOv4\_AF\cite{zhao2022improved}}, YOLOv5s \cite{jocher2020ultralytics}, YOLOv5s with attention feature fusion block (AFFB-YOLOv5s) \cite{lian2021small}, \textcolor{black}{YOLOv5 with ACmix attention mechanism (YOLOv5\_ACmix)\cite{he2023vehicle}, Dynamic RegionAware Convolution (DRConv)\cite{chen2021dynamic}, Dynamic convolution (DC)\cite{chen2020dynamic}}, Densenet \cite{iandola2014densenet}, RetinaNet \cite{lin2017focal}, EfficientNet \cite{yang2017pannet}, YOLOv7 \cite{wang2023yolov7} and cross-scale and illumination-invariant detection model (CSIM) \cite{lu2023cross} on the BDD100K dataset, UA-DETRAC dataset and BIT-Vehicle dataset to compute their mAP values to test their performance for vehicle detection as shown TABLEs II, III and IV. As shown in TABLE II, we can see that our DTNet has obtained an improvement of 4.6\% than the second detection method Modified YOLOv4 on the BDD100K dataset in terms of mAP, which shows effectiveness of the proposed DTNet for vehicle detection. As listed in TABLE III, we can see that our DTNet has a higher mAP value of 1.3\% than that of the second DC and outperforms the third DRConv by 1.6\% in terms of mAP value on UA-DETRAC dataset for vehicle detection. \textcolor{black}{We can see that the proposed DTNet has also obtained higher mAP than that of YOLOv7 and CSIM, which shows that the proposed DTNet is more effective than that state-of-the-art methods in the filed of dynamic convolutions.} As illustrated in TABLE IV, we can see that our DTNet has obtained the best detection result for vehicle detection on the BIT-Vehicle dataset for vehicle detection. For instance, DTNet exceeds the second YOLOv3 by 3.4\% in terms of mAP. \textcolor{black}{Furthermore, our DTNet has obtained better detection result than that of recent YOLOv5s, which shows effectiveness of the proposed DTNet for vehicle detection.}  \textcolor{black}{To comprehensively test performance of our proposed method, we use Precision, Recall and mAP with varying thresholds as metrics to conduct experiments, where mAP with varying thresholds is an average value of mAP values with varying thresholds from 0.5 to 0.95 by increasing of 0.05 while the mAP is set to threshold of 0.5 in TABLEs I-V. As shown in TABLE V, we can see that our method has obtained the best performance than that of YOLOv5s and YOLOv7 in terms of Percision, Recall, mAP and mAP with varying thresholds, which shows effectiveness of the proposed method UA-DETRAC dataset.} 

Qualitative analysis uses different methods, i.e., \textcolor{black}{YOLOv5s}, YOLOv7, and our DTNet on a scene to accurately detect more vehicles to test their performance in vehicle detection. As shown in Fig.2, we can see that our method 
better detects a car behind a bus station than those of YOLOv5s and YOLOv7. \textcolor{black}{Specifically, DTNet detected occluded vehicles, while other methods failed to detect the occluded ones (missed vehicles are highlighted with yellow dashed lines).} As shown in Fig.3, we can see that YOLOv5s cannot detect \textcolor{black}{occluded vehicles and smaller-sized vehicles} in the upper left corner of Fig.3. Also, YOLOv7 erroneously recognizes people as cars. \textcolor{black}{False positives and false negatives have been highlighted with yellow dashed lines.} As shown in Fig.4, we can see that YOLOv5s and YOLOv7 cannot detect some cars \textcolor{black}{due to lighting and occlusion issues, as indicated by yellow dashed lines in Fig.4. \textcolor{black}{thus, our DTNet is effective in occlusion, vehicle size, and lighting conditions for vehicle detection.} \textcolor{black}{To further verify  performance of the proposed method in vehicle detection method, precision-recall curve is used to conduct experiments. That is, the larger area enclosed by the curve shows better performance of vehicle detection method. In Fig.5, we can see that our method surpasses YOLOv7 in terms of precision-recall curve.} And the proposed DTNet is effective for vehicle detection in terms of qualitative analysis. In summary, our DTNet is competitive for vehicle detection. Due to effects of varying environments, i.e., sunny and snowy, performance of the proposed DTNet may decrease. In the future, we will combine signal processing knowledge, discriminative learning and video processing techniques to design deep networks to improve robustness for vehicle detection under bad weather.} 

\section{Conclusion}
In this paper, we present a dynamic Transformer network for vehicle detection. This paper implements a dynamic adaptive detector by a dynamic convolution, according to different traffic scenes. Taking into relations of different information account, a mixed attention mechanism fused channel attention and Transformer to strengthen effects of channels and pixels to extract more salient information for vehicle detection. To overcome native effect of differences within an image, a translation variant convolution utilizes spatial location information to guide a CNN to extract more accurate information for vehicle detection. Our model has achieved better performance for vehicle detection. In the future, we will conduct research in a self-supervised learning approach to further enhance the robustness and accuracy of vehicle detection under varying environmental conditions. 




\bibliographystyle{IEEEtran}
\bibliography{ref}

\begin{thebibliography}{10}
\providecommand{\url}[1]{#1}
\csname url@samestyle\endcsname
\providecommand{\newblock}{\relax}
\providecommand{\bibinfo}[2]{#2}
\providecommand{\BIBentrySTDinterwordspacing}{\spaceskip=0pt\relax}
\providecommand{\BIBentryALTinterwordstretchfactor}{4}
\providecommand{\BIBentryALTinterwordspacing}{\spaceskip=\fontdimen2\font plus
\BIBentryALTinterwordstretchfactor\fontdimen3\font minus
  \fontdimen4\font\relax}
\providecommand{\BIBforeignlanguage}[2]{{%
\expandafter\ifx\csname l@#1\endcsname\relax
\typeout{** WARNING: IEEEtran.bst: No hyphenation pattern has been}%
\typeout{** loaded for the language `#1'. Using the pattern for}%
\typeout{** the default language instead.}%
\else
\language=\csname l@#1\endcsname
\fi
#2}}
\providecommand{\BIBdecl}{\relax}
\BIBdecl

\bibitem{singh2024smart}
A.~Singh, M.~Z.~U. Rahma, P.~Rani, R.~Sharma, E.~Kariri \emph{et~al.}, ``Smart
  traffic monitoring through real-time moving vehicle detection using deep
  learning via aerial images for consumer application,'' \emph{IEEE
  Transactions on Consumer Electronics}, vol.~70, no.~4, pp. 7302--7309, 2024.

\bibitem{gao2024monoli}
H.~Gao, X.~Yu, Y.~Xu, J.~Y. Kim, and Y.~Wang, ``Monoli: Precise monocular 3-d
  object detection for next-generation consumer electronics for autonomous
  electric vehicles,'' \emph{IEEE Transactions on Consumer Electronics},
  vol.~70, no.~1, pp. 3475--3486, 2024.

\bibitem{dalal2005histograms}
N.~Dalal and B.~Triggs, ``Histograms of oriented gradients for human
  detection,'' in \emph{2005 IEEE Computer Society Conference on Computer
  Vision and Pattern Recognition}, vol.~1.\hskip 1em plus 0.5em minus
  0.4em\relax IEEE, 2005, pp. 886--893.

\bibitem{chen2012vehicle}
Z.~Chen, T.~Ellis, and S.~A. Velastin, ``Vehicle detection, tracking and
  classification in urban traffic,'' in \emph{2012 15th International IEEE
  Conference on Intelligent Transportation Systems}.\hskip 1em plus 0.5em minus
  0.4em\relax IEEE, 2012, pp. 951--956.

\bibitem{minaeian2015vision}
S.~Minaeian, J.~Liu, and Y.-J. Son, ``Vision-based target detection and
  localization via a team of cooperative uav and ugvs,'' \emph{IEEE
  Transactions on Systems, Man, and Cybernetics: Systems}, vol.~46, no.~7, pp.
  1005--1016, 2015.

\bibitem{karaimer2017detection}
H.~C. Karaimer, I.~Baris, and Y.~Bastanlar, ``Detection and classification of
  vehicles from omnidirectional videos using multiple silhouettes,''
  \emph{Pattern Analysis and Applications}, vol.~20, pp. 893--905, 2017.

\bibitem{chang2009online}
W.-C. Chang and C.-W. Cho, ``Online boosting for vehicle detection,''
  \emph{IEEE Transactions on Systems, Man, and Cybernetics, Part B
  (Cybernetics)}, vol.~40, no.~3, pp. 892--902, 2009.

\bibitem{zhou2007moving}
J.~Zhou, D.~Gao, and D.~Zhang, ``Moving vehicle detection for automatic traffic
  monitoring,'' \emph{IEEE Transactions on Vehicular Technology}, vol.~56,
  no.~1, pp. 51--59, 2007.

\bibitem{lim2017real}
K.~Lim, Y.~Hong, Y.~Choi, and H.~Byun, ``Real-time traffic sign recognition
  based on a general purpose gpu and deep-learning,'' \emph{PLoS one}, vol.~12,
  no.~3, p. e0173317, 2017.

\bibitem{duan2019centernet}
K.~Duan, S.~Bai, L.~Xie, H.~Qi, Q.~Huang, and Q.~Tian, ``Centernet: Keypoint
  triplets for object detection,'' in \emph{Proceedings of the IEEE/CVF
  International Conference On Computer Vision}, 2019, pp. 6569--6578.

\bibitem{rani2021littleyolo}
E.~Rani \emph{et~al.}, ``Littleyolo-spp: A delicate real-time vehicle detection
  algorithm,'' \emph{Optik}, vol. 225, p. 165818, 2021.

\bibitem{hu2018sinet}
X.~Hu, X.~Xu, Y.~Xiao, H.~Chen, S.~He, J.~Qin, and P.-A. Heng, ``Sinet: A
  scale-insensitive convolutional neural network for fast vehicle detection,''
  \emph{IEEE Transactions on Intelligent Transportation Systems}, vol.~20,
  no.~3, pp. 1010--1019, 2018.

\bibitem{song2019vision}
H.~Song, H.~Liang, H.~Li, Z.~Dai, and X.~Yun, ``Vision-based vehicle detection
  and counting system using deep learning in highway scenes,'' \emph{European
  Transport Research Review}, vol.~11, no.~1, pp. 1--16, 2019.

\bibitem{dong2022lightweight}
X.~Dong, S.~Yan, and C.~Duan, ``A lightweight vehicles detection network model
  based on yolov5,'' \emph{Engineering Applications of Artificial
  Intelligence}, vol. 113, p. 104914, 2022.

\bibitem{roy2022multi}
D.~Roy, Y.~Li, T.~Jian, P.~Tian, K.~R. Chowdhury, and S.~Ioannidis,
  ``Multi-modality sensing and data fusion for multi-vehicle detection,''
  \emph{IEEE Transactions on Multimedia}, 2022.

\bibitem{zhang2022real}
Y.~Zhang, Z.~Guo, J.~Wu, Y.~Tian, H.~Tang, and X.~Guo, ``Real-time vehicle
  detection based on improved yolo v5,'' \emph{Sustainability}, vol.~14,
  no.~19, p. 12274, 2022.

\bibitem{mo2020improved}
N.~Mo and L.~Yan, ``Improved faster rcnn based on feature amplification and
  oversampling data augmentation for oriented vehicle detection in aerial
  images,'' \emph{Remote Sensing}, vol.~12, no.~16, p. 2558, 2020.

\bibitem{manana2018preprocessed}
M.~Manana, C.~Tu, and P.~A. Owolawi, ``Preprocessed faster rcnn for vehicle
  detection,'' in \emph{2018 International Conference on Intelligent and
  Innovative Computing Applications}.\hskip 1em plus 0.5em minus 0.4em\relax
  IEEE, 2018, pp. 1--4.

\bibitem{balamuralidhar2021multeye}
N.~Balamuralidhar, S.~Tilon, and F.~Nex, ``Multeye: Monitoring system for
  real-time vehicle detection, tracking and speed estimation from uav imagery
  on edge-computing platforms,'' \emph{Remote sensing}, vol.~13, no.~4, p. 573,
  2021.

\bibitem{kim2018performance}
K.-J. Kim, P.-K. Kim, Y.-S. Chung, and D.-H. Choi, ``Performance enhancement of
  yolov3 by adding prediction layers with spatial pyramid pooling for vehicle
  detection,'' in \emph{2018 15th IEEE International Conference on Advanced
  Video and Signal Based Surveillance}.\hskip 1em plus 0.5em minus 0.4em\relax
  IEEE, 2018, pp. 1--6.

\bibitem{kumar2022deep}
S.~Kumar, A.~Jain, S.~Rani, H.~Alshazly, S.~A. Idris, and S.~Bourouis, ``Deep
  neural network based vehicle detection and classification of aerial images.''
  \emph{Intelligent Automation \& Soft Computing}, vol.~34, no.~1, 2022.

\bibitem{humayun2022traffic}
M.~Humayun, F.~Ashfaq, N.~Z. Jhanjhi, and M.~K. Alsadun, ``Traffic management:
  Multi-scale vehicle detection in varying weather conditions using yolov4 and
  spatial pyramid pooling network,'' \emph{Electronics}, vol.~11, no.~17, p.
  2748, 2022.

\bibitem{charouh2022resource}
Z.~Charouh, A.~Ezzouhri, M.~Ghogho, and Z.~Guennoun, ``A resource-efficient
  cnn-based method for moving vehicle detection,'' \emph{Sensors}, vol.~22,
  no.~3, p. 1193, 2022.

\bibitem{chen2020dynamic}
Y.~Chen, X.~Dai, M.~Liu, D.~Chen, L.~Yuan, and Z.~Liu, ``Dynamic convolution:
  Attention over convolution kernels,'' in \emph{Proceedings of the IEEE/CVF
  conference on Computer Vision and Pattern Recognition}.\hskip 1em plus 0.5em
  minus 0.4em\relax IEEE, 2020, pp. 11\,030--11\,039.

\bibitem{verelst2020dynamic}
T.~Verelst and T.~Tuytelaars, ``Dynamic convolutions: Exploiting spatial
  sparsity for faster inference,'' in \emph{Proceedings of the IEEE/CVF
  Conference on Computer Vision and Pattern Recognition}.\hskip 1em plus 0.5em
  minus 0.4em\relax IEEE, 2020, pp. 2320--2329.

\bibitem{chen2021dynamic}
J.~Chen, X.~Wang, Z.~Guo, X.~Zhang, and J.~Sun, ``Dynamic region-aware
  convolution,'' in \emph{Proceedings of the IEEE/CVF Conference on Computer
  Vision and Pattern Recognition}.\hskip 1em plus 0.5em minus 0.4em\relax IEEE,
  2021, pp. 8064--8073.

\bibitem{hu2018squeeze}
J.~Hu, L.~Shen, and G.~Sun, ``Squeeze-and-excitation networks,'' in
  \emph{Proceedings of the IEEE Conference on Computer Vision and Pattern
  Recognition}.\hskip 1em plus 0.5em minus 0.4em\relax IEEE, 2018, pp.
  7132--7141.

\bibitem{zhang2020residual}
Y.~Zhang, Y.~Tian, Y.~Kong, B.~Zhong, and Y.~Fu, ``Residual dense network for
  image restoration,'' \emph{IEEE Transactions on Pattern Analysis and Machine
  Intelligence}, vol.~43, no.~7, pp. 2480--2495, 2020.

\bibitem{wang2023yolov7}
C.-Y. Wang, A.~Bochkovskiy, and H.-Y.~M. Liao, ``Yolov7: Trainable
  bag-of-freebies sets new state-of-the-art for real-time object detectors,''
  in \emph{Proceedings of the IEEE/CVF Conference on Computer Vision and
  Pattern Recognition}.\hskip 1em plus 0.5em minus 0.4em\relax IEEE, 2023, pp.
  7464--7475.

\bibitem{shen2023adaptive}
H.~Shen, Z.-Q. Zhao, and W.~Zhang, ``Adaptive dynamic filtering network for
  image denoising,'' in \emph{Proceedings of the AAAI Conference on Artificial
  Intelligence}, vol.~37, no.~2.\hskip 1em plus 0.5em minus 0.4em\relax AAAI
  Press, 2023, pp. 2227--2235.

\bibitem{ioffe2015batch}
S.~Ioffe and C.~Szegedy, ``Batch normalization: Accelerating deep network
  training by reducing internal covariate shift,'' in \emph{International
  Conference on Machine Learning}.\hskip 1em plus 0.5em minus 0.4em\relax pmlr,
  2015, pp. 448--456.

\bibitem{elfwing2018sigmoid}
S.~Elfwing, E.~Uchibe, and K.~Doya, ``Sigmoid-weighted linear units for neural
  network function approximation in reinforcement learning,'' \emph{Neural
  Networks}, vol. 107, pp. 3--11, 2018.

\bibitem{chen2023activating}
X.~Chen, X.~Wang, J.~Zhou, Y.~Qiao, and C.~Dong, ``Activating more pixels in
  image super-resolution transformer,'' in \emph{Proceedings of the IEEE/CVF
  Conference on Computer Vision and Pattern Recognition}.\hskip 1em plus 0.5em
  minus 0.4em\relax IEEE, 2023, pp. 22\,367--22\,377.

\bibitem{chen2022tvconv}
J.~Chen, T.~He, W.~Zhuo, L.~Ma, S.~Ha, and S.-H.~G. Chan, ``Tvconv: Efficient
  translation variant convolution for layout-aware visual processing,'' in
  \emph{Proceedings of the IEEE/CVF Conference on Computer Vision and Pattern
  Recognition}.\hskip 1em plus 0.5em minus 0.4em\relax IEEE, 2022, pp.
  12\,548--12\,558.

\bibitem{wen2020ua}
L.~Wen, D.~Du, Z.~Cai, Z.~Lei, M.-C. Chang, H.~Qi, J.~Lim, M.-H. Yang, and
  S.~Lyu, ``Ua-detrac: A new benchmark and protocol for multi-object detection
  and tracking,'' \emph{Computer Vision and Image Understanding}, vol. 193, p.
  102907, 2020.

\bibitem{yu2020bdd100k}
F.~Yu, H.~Chen, X.~Wang, W.~Xian, Y.~Chen, F.~Liu, V.~Madhavan, and T.~Darrell,
  ``Bdd100k: A diverse driving dataset for heterogeneous multitask learning,''
  in \emph{Proceedings of the IEEE/CVF Conference on Computer Vision and
  Pattern Recognition}.\hskip 1em plus 0.5em minus 0.4em\relax IEEE, 2020, pp.
  2636--2645.

\bibitem{dong2015vehicle}
Z.~Dong, Y.~Wu, M.~Pei, and Y.~Jia, ``Vehicle type classification using a
  semisupervised convolutional neural network,'' \emph{IEEE Transactions on
  Intelligent Transportation Systems}, vol.~16, no.~4, pp. 2247--2256, 2015.

\bibitem{bottou2012stochastic}
L.~Bottou, ``Stochastic gradient descent tricks,'' in \emph{Neural Networks:
  Tricks of the Trade: Second Edition}.\hskip 1em plus 0.5em minus 0.4em\relax
  Springer, 2012, pp. 421--436.

\bibitem{stevens2020deep}
E.~Stevens, L.~Antiga, and T.~Viehmann, \emph{Deep Learning with PyTorch:
  Build, train, and tune neural networks using Python tools}.\hskip 1em plus
  0.5em minus 0.4em\relax Manning, 2020.

\bibitem{redmon2018yolov3}
J.~Redmon and A.~Farhadi, ``Yolov3: An incremental improvement,'' \emph{arXiv
  preprint arXiv:1804.02767}, 2018.

\bibitem{liu2016ssd}
W.~Liu, D.~Anguelov, D.~Erhan, C.~Szegedy, S.~Reed, C.-Y. Fu, and A.~C. Berg,
  ``Ssd: Single shot multibox detector,'' in \emph{in Proceedings of European
  Conference on Computer Vision}.\hskip 1em plus 0.5em minus 0.4em\relax
  Springer, 2016, pp. 21--37.

\bibitem{han2020wasserstein}
Y.~Han, X.~Liu, Z.~Sheng, Y.~Ren, X.~Han, J.~You, R.~Liu, and Z.~Luo,
  ``Wasserstein loss-based deep object detection,'' in \emph{Proceedings of the
  IEEE/CVF Conference on Computer Vision and Pattern Recognition
  Workshops}.\hskip 1em plus 0.5em minus 0.4em\relax IEEE, 2020, pp. 998--999.

\bibitem{bochkovskiy2020yolov4}
A.~Bochkovskiy, C.-Y. Wang, and H.-Y.~M. Liao, ``Yolov4: Optimal speed and
  accuracy of object detection,'' \emph{arXiv preprint arXiv:2004.10934}, 2020.

\bibitem{li2020deep}
Y.~Li, H.~Wang, L.~M. Dang, T.~N. Nguyen, D.~Han, A.~Lee, I.~Jang, and H.~Moon,
  ``A deep learning-based hybrid framework for object detection and recognition
  in autonomous driving,'' \emph{IEEE Access}, vol.~8, pp. 194\,228--194\,239,
  2020.

\bibitem{zhao2022improved}
J.~Zhao, S.~Hao, C.~Dai, H.~Zhang, L.~Zhao, Z.~Ji, and I.~Ganchev, ``Improved
  vision-based vehicle detection and classification by optimized yolov4,''
  \emph{IEEE Access}, vol.~10, pp. 8590--8603, 2022.

\bibitem{jocher2020ultralytics}
G.~Jocher, A.~Stoken, J.~Borovec, L.~Changyu, A.~Hogan, L.~Diaconu,
  J.~Poznanski, L.~Yu, P.~Rai, R.~Ferriday \emph{et~al.}, ``ultralytics/yolov5:
  v3. 0,'' \emph{Zenodo}, 2020.

\bibitem{lian2021small}
J.~Lian, Y.~Yin, L.~Li, Z.~Wang, and Y.~Zhou, ``Small object detection in
  traffic scenes based on attention feature fusion,'' \emph{Sensors}, vol.~21,
  no.~9, p. 3031, 2021.

\bibitem{he2023vehicle}
X.~He, ``Vehicle target detection algorithm based on yolov5,'' \emph{Frontiers
  in Computing and Intelligent Systems}, vol.~3, no.~1, pp. 56--59, 2023.

\bibitem{iandola2014densenet}
F.~Iandola, M.~Moskewicz, S.~Karayev, R.~Girshick, T.~Darrell, and K.~Keutzer,
  ``Densenet: Implementing efficient convnet descriptor pyramids,'' \emph{arXiv
  preprint arXiv:1404.1869}, 2014.

\bibitem{lin2017focal}
T.-Y. Lin, P.~Goyal, R.~Girshick, K.~He, and P.~Doll{\'a}r, ``Focal loss for
  dense object detection,'' in \emph{Proceedings of the IEEE International
  Conference on Computer Vision}, 2017, pp. 2980--2988.

\bibitem{yang2017pannet}
J.~Yang, X.~Fu, Y.~Hu, Y.~Huang, X.~Ding, and J.~Paisley, ``Pannet: A deep
  network architecture for pan-sharpening,'' in \emph{Proceedings of the IEEE
  International Conference on Computer Vision}.\hskip 1em plus 0.5em minus
  0.4em\relax IEEE, 2017, pp. 5449--5457.

\bibitem{lu2023cross}
Y.-F. Lu, J.-W. Gao, Q.~Yu, Y.~Li, Y.-S. Lv, and H.~Qiao, ``A cross-scale and
  illumination invariance-based model for robust object detection in traffic
  surveillance scenarios,'' \emph{IEEE Transactions on Intelligent
  Transportation Systems}, 2023.

\end{thebibliography}

\end{document}